# Improving Arabic Diacritization by Learning to Diacritize and Translate

**Brian Thompson**
Apple
brian_thompson2@apple.com

**Ali Alshehri**
Apple
a_alshehri@apple.com

## Abstract

We propose a novel multitask learning method for diacritization which trains a model to both diacritize and translate. Our method addresses data sparsity by exploiting large, readily available bitext corpora. Furthermore, translation requires implicit linguistic and semantic knowledge, which is helpful for resolving ambiguities in the diacritization task. We apply our method to the Penn Arabic Treebank and report a new state-of-the-art word error rate of 4.79%. We also conduct manual and automatic analysis to better understand our method and highlight some of the remaining challenges in diacritization.

هيا لنذهب → هَيَّا لِنَذْهَبْ

[hjaː lnðhb] [hajːaː linaðhab]

Figure 1: Arabic diacritization is the task of adding diacritics (markings above and below characters, shown in red) to Arabic text. Diacritics clarify how a word is pronounced, including short vowels and elongation, and disambiguate word meaning. Here, we show the diacritization of هيا لنذهب (let's go). The IPA pronunciations below each word demonstrate that the diacritics are crucial for pronouncing each word: the undiacritized form maps to an incorrect pronunciation, while the diacritized form maps to the correct pronunciation (the contributions the diacritics make to the pronunciation are also shown in red).

## 1 Introduction

Arabic is typically written without short vowels and other pronunciation indication markers, collectively referred to as diacritics.[1] A longstanding task in Natural Language Processing (NLP) is to take undiacritized text and add the diacritics, referred to as diacritization (see Figure 1). Diacritics not only indicate how to pronounce a word; they also resolve ambiguities in meaning between different words with the same (undiacritized) written form.

Diacritization has applications in Automatic Speech Recognition (ASR; Vergyri and Kirchhoff, 2004; Ananthakrishnan et al., 2005; Biadsy et al., 2009), Machine Translation (MT; Diab et al., 2007) morphological analysis (Habash et al., 2016), Arabic lexical recognition tests (Hamed and Zesch, 2018; Hamed, 2019), and homograph resolution (Alqahtani et al., 2019a).

Diacritization is the dominant source of errors in Arabic grapheme to phoneme conversion (Ali et al., 2020), a crucial component in Text-to-Speech (TTS). With the rise of personal digital assistants with TTS capabilities, there is a clear need for improved automatic diacritization methods.

We focus on Modern Standard Arabic (MSA), a standardized dialect of Arabic used in most academic, legal, and news publications, and an obvious choice for TTS systems. MSA is the 5th most spoken language in the world with about 274M speakers[2] (Eberhard et al., 2021).

### 1.1 Challenge #1: Data Sparsity

Arabic is a morphologically rich language, where significant information concerning syntactic units and relations is expressed at word-level. For example, a word like فأسقيناكموه is roughly translated to: 'and we gave it to you to drink.' This fact results in Arabic having a large vocabulary (by way of example, the number of unique, undiacritized words in the Ara-bic bible from Christodouloupoulos and Steedman (2015) is about 4.38x larger than the number of unique, lower-cased words in the English equivalent.) Finally, high-quality diacritized datasets tend to be quite small: The Penn Arabic Treebank (PATB) training subset used in this work is only 15,789 lines, and data available in other dialects can be substantially smaller. These factors result in diacritized Arabic being very data sparse, with diacritics models typically needing to produce a large number of unseen words.

[1] Notable exceptions include the Quran and many children's books.

[2] "Speaker" is a bit of a misnomer: Most Arabic speakers can understand MSA but would not typically produce it.

### 1.2 Challenge #2: Ambiguity

Arabic can express gender (male, female), number (singular, dual, plural), case (nominative, accusative, genitive), aspect (perfect, imperfect), voice (active, passive) and mood (indicative, imperative, subjunctive) at the word level, and distinctions are often conveyed using only diacritics. This results in undiacritized Arabic having a huge number of homographs; Debili et al. (2002) report an average of 11.6 possible diacritizations for each undiacritized word in Arabic. These homographs must be resolved by a diacritics model in order to correctly diacritize Arabic text.

### 1.3 Overview of Proposed Method

We propose a novel Multitask Learning (MTL; Caruana, 1997) based approach to diacritization. Specifically, we propose augmenting diacritics training data with bitext in order to to train a model to both diacritize Arabic and translate into and out of Arabic.

Our approach addresses data sparsity by substantially increasing the amount of training data seen by the model, since it enables the use of large, readily available MT datasets (i.e. bitext). In our experiments on the PATB, adding bitext increases training data from 502k to 138M (non-unique) Arabic words, and decreases the Out-of-Vocabulary (OOV) rate from 7.33% to 1.14%. In contrast, prior MTL work in diacritization has used hand-curated features such as parts of speech, gender, and case (see §2.1), severely limiting both the amount of available data and the applicability to languages without such resources.

Our approach also addresses ambiguity, since task of translation requires (implicit) semantic and linguistic knowledge. Training on bitext injects semantic and linguistic knowledge into the model which is helpful for resolving ambiguities in diacritization (see Table 1).

These factors contribute to our method achieving a new State-of-the-Art (SOTA) Word Error Rate (WER) of 4.79% on the PATB, vs 7.49% for an equivalent baseline without MTL.

### 1.4 Main Contributions of This Work

The main contributions of this work are:

- We present a novel MTL approach for MSA diacritization, which does not require a morphological analyzer or specialized annotations, and thus is likely extensible to other languages and domains.
- We achieve a new SOTA WER of 4.79% on the PATB test set.
- We perform extensive automatic analysis of our method to see how it performs on various conditions including different parts of speech, genders, word frequencies, and sentence lengths.
- We perform detailed manual error analysis of our method, illustrating both issues in the PATB dataset as well as the remaining challenges in Arabic diacritization.

## 2 Related Word

### 2.1 Diacritization

Many works have explored using neural networks for Arabic diacritization (Zalmout and Habash, 2017, 2019; Alqahtani and Diab, 2019; Alqahtani et al., 2019b).

Alqahtani et al. (2020) and Zalmout and Habash (2020) both explored MTL regimes in which a model learns to predict Arabic diacritics simultaneously with other features in the PATB. Alqahtani et al. (2020) used additional features of syntactic diacritization, word segmentation, and Part of Speech (POS) tagging, and achieved a WER of 7.51%,[3] while Zalmout and Habash (2020) used additional features of lemmas, aspect, case, gender, person, POS, number, mood, state, voice, enclitics, and proclitics, and a achieved a WER of 7.2%. By also adding an external morphological analyzer, they improved WER to 6.1%.

[3] Unless stated otherwise, all word error rates in this work correspond to test set from the PATB data divisions proposed by Diab et al. (2013).

| # | Arabic Sentence | English Sentence | Voiced | Pronunciation | Translation |
|---|---|---|---|---|---|
| 0 | **علم** السعودية أخضر وأبيض اللون | The **flag** of Saudi Arabia is green and white | عَلَمُ | [ʕalamu] | flag |
| 1 | أحب **علم** الفلك | I love space **science** | عِلمَ | [ʕilma] | science |
| 2 | **علم** ناصر أحمد السباحة | Nasser **taught** Ahmad how to swim | عَلَّمَ | [ʕalːama] | taught |

Table 1: Adding bitext to our training data improves the semantic and linguistic knowledge of our diacritization model. For example, in order to correctly translate علم out of Arabic, the model must learn to implicitly perform homographic resolution to determine if the word is being used to mean "flag," "science," "taught," or other meanings. This knowledge is helpful for diacritization since diacritized forms are intrinsically linked with word meaning. The model can also implicitly learn, for example, that علم in example #2 is being used as a causative past tense verb. This can help the model diacritize this use of علم correctly (عَلَّمَ), even if عَلَّمَ does not appear in the diacritization training data, since عَلَّمَ follows a common diacritization pattern for causative past tense verbs.

These works illustrate the potential of MTL, but they require additional hand-curated features. This limits both the datasets they can use (neither are able to take advantage of large additional datasets) and the languages they could be applied to.

#### 2.1.1 Contextual Embeddings

Náplava et al. (2021) showed that contextual embeddings can result in substantial improvements in diacritization error rates in several languages, but unfortunately they did not report results on Arabic.

Qin et al. (2021) started with a strong baseline built on ZEN 2.0 (Song et al., 2021), an n-gram aware BERT variant. Their BERT-based baseline outperformed prior work on PATB. They then claimed even stronger results on PATB with two methods that incorporate multitask training with a second, auxiliary decoder trained to predict the diacritics produced by the Farasa morphological analyzer (Abdelali et al., 2016). We argue that their experimental setup was fundamentally flawed, since Farasa was trained on the PATB test set[4] and can leak information about the test set to the model.[5] They also reported results on the Tashkeela training/test data (Zerrouki and Balla, 2017; Fadel et al., 2019), which does not have a potential testset contamination problem, and found that their method under-performs a straightforward bidirectional LSTM,[6] which supports the hypothesis that their strong PATB results are due to training on a derivative of the test set.

### 2.2 Character-level, Multilingual MT

Multilingual MT (Dong et al., 2015) has been shown to dramatically improve low-resource translation, including enabling transfer from higher resource language pairs to lower-resource language pairs (Zoph et al., 2016; Nguyen and Chiang, 2017; Neubig and Hu, 2018). In contrast, we encourage transfer from undiacritized Arabic to much lower-resourced diacritized Arabic.

Most MT systems operate at the subword level (Sennrich et al., 2016; Kudo and Richardson, 2018); however, such approaches could result in a diacritized word having little or no subword overlap with the undiacritized form of the same word. We instead train a character-level encoder-decoder model (Lee et al., 2017; Cherry et al., 2018), which has been shown to outperform subword-level models at diacritization (Alqahtani and Diab, 2019), to maximize the amount of shared representation between diacritized and undiacritized words.

## 3 Experiments

We train a character-level transformer encoder-decoder model on both the diacritics data and bitext. Our primary model performs diacriti-zation; translation from Arabic (Ar) to English (En), French (Fr), and Spanish (Es); and translation from English, French, and Spanish to Arabic. We also perform ablations for analysis purposes, leaving out (1) the Ar→∗ data, (2) the ∗→Ar data, and (3) all of the bitext.

[4] Farasa was trained on PATB parts 1, 2 and 3 *in their entirety*, and then tested on a separate collection of hand curated news articles (Abdelali et al., 2016)

[5] To understand how leakage from the test set can occur, consider the word النجمة (the star; female). النجمة appears three times in the training data, once without diacritics (likely an error) and twice as النَّجْمَةِ (the star; genitive case). However, it appears 9 times in the test set, each time diacritized as النَّجْمَة (the star; without case ending). Farasa is trained on both the training and test data, so from Farasa's perspective, النَّجْمَة is by far the most likely diacritization of النجمة . Thus when the model sees النجمة in training, Farasa can artificially bias the model toward producing the diacritized form in the test set, despite that form never appearing in the training data.

[6] Qin et al. (2021) claimed to achieve state-of-the-art performance on both datasets, but this is not supported by their results (see their Table 2, noting that bold does not denote the best system).

| Name | Form | Sound [IPA] |
|---|---|---|
| Fatha | ◌َ | /a/ |
| Fathatan | ◌ً | /an/ |
| Kasra | ◌ِ | /i/ |
| Kasratan | ◌ٍ | /in/ |
| Damma | ◌ُ | /u/ |
| Dammatan | ◌ٌ | /un/ |
| Dagger Alif | ◌ٰ | /aː/ |
| Maddah | ◌ٓ | /ʕaː/ |
| Shadda | ◌ّ | Elongation (ː) |
| Sukun | ◌ْ | None |

Table 2: Diacritics considered in this work

| | Ar-En | Ar-Es | Ar-Fr | Diacs |
|---|---|---|---|---|
| Global Voices | 0.9 | 0.9 | 0.5 | - |
| CCAligned | - | 21.9 | 21.7 | - |
| News Commentary | 5.0 | 5.0 | 4.3 | - |
| United Nations | 20.7 | 19.9 | 19.5 | - |
| WikiMatrix | 15.0 | 1.7 | 1.6 | - |
| PATB | - | - | - | 0.5 |
| Total | 40.8 | 48.4 | 47.1 | 0.5 |

Table 3: Size (millions of Arabic words) of training datasets used in this work. Note that total bitext is about 275x larger than diacritics data.

Each model uses a single encoder and decoder for all tasks. During training, we prepend a tag to each output sentence to tell the model whether the output is undiacritized Arabic, diacritized Arabic, English, French, or Spanish. At inference time we force decode the tag to request that the model produce diacritized Arabic.

### 3.1 Decoding

During decoding, we keep track of which input characters the decoder has produced and constrain the decoder as follows: If the previous output is a non-diacritic Arabic character, we restrict the decoder to produce any diacritic or the next input character. If the previous output is a shadda, we restrict the decoder to produce a non-shadda diacritic or the next input character. Otherwise, the model is forced to produce the next input character. Without these restrictions, we found that the model would occasionally produce minor paraphrastic variations of the input.[7]

### 3.2 Diacritics Data

We use PATB part 1 v4.1 (LDC2010T13), part 2 v3.1 (LDC2011T09) and part 3 v3.2 (LDC2010T08), following the train/dev/test splits proposed by Diab et al. (2013). We perform unicode NFKD normalization on the text in order to (1) split Unicode characters which contain both a non-diacritic and diacritic (e.g. the Unicode character for alif with maddah above (U+0622) is split into alif (U+0627) and maddah (U+0653)) and (2) normalize the order of characters (e.g. alif + high hamza + fatha and alif + fatha + high hamza both render as أَ and are normalized to alif + high hamza + fatha). The diacritics considered in this work are shown in Table 2.

### 3.3 MT Data

We use Ar↔{En,Fr,Es} data from Wikimatrix (Schwenk et al., 2019), Global Voices,[8] United Nations (Ziemski et al., 2016), and News-Commentary,[9] and Ar↔{Fr,Es} data from CCAligned (El-Kishky et al., 2020), after joining on English urls. We filter out noisy sentence pairs using the scripts[10] provided by Thompson and Post (2020) using more aggressive thresholds of min_laser_score=1.06, max_3gram_overlap=0.1 for the CCAligned data and using values from Thompson and Post (2020) otherwise. We limit each dataset to 1M lines per language pair, so that no single data source dominates training. Data size are shown in Table 3. We up-sample PATB by 20x when combining it with the bitext, since it is much smaller than the bitext.

Adding bitext significantly reduces the OOV rates of the model: see Table 4.

### 3.4 Long Sentence Handling

The computational complexity of Transformer layers is proportional to sequence length squared (Vaswani et al., 2017), so we do not want to train or run inference on arbitrarily long sequences of characters. Instead, we limit the maximum input and output sequence length to 600.

[7] The tendency of a multilingual MT models to paraphrase input has been noted (and exploited) by Tiedemann and Scherrer (2019) and Thompson and Post (2020).

[8] casmacat.eu/corpus/global-voices.html

[9] data.statmt.org/news-commentary/

[10] github.com/thompsonb/prism_bitext_filter

| Training Data | OOV Rate | |
|---|---|---|
| | Undiacritized | Diacritized |
| PATB | 7.33% | 10.54% |
| PATB + Bitext | 1.14% | 9.56% |

Table 4: OOV rates (rate of seeing word in test that was not seen in train), for encoder (which sees words without diacritics) and decoder (which produces words with diacritics). The bitext brings down the OOV rate substantially for the Encoder. We were surprised that the decoder OOV rate wend down by adding (undiacritized) bitext; manual inspection showed this was because some a small percentage words in PATB are missing diacritics (see also: §6).

To diacritize a sentence with more than 300 input characters, we take overlapping windows of 300 characters with a step size of 100 characters. We predict diacritics independently for each window, and construct the diacritized output sentence using portions of each window that have at least 100 characters of context on the input. For the bitext data, we simply discard sentence pairs with greater than 600 input or output characters.

### 3.5 Models & Training

We train character-level Transformer (Vaswani et al., 2017) models in fairseq (Ott et al., 2019). Hyperparameters are tuned on the development set.

The (non-MTL) baseline has 6 encoder and decoder layers, encoder and decoder embedding dimensions of 1024, encoder and decoder feed-forward network embedding dimensions of 8192, and 16 heads. All embeddings are shared. We train with a learning rate of 0.0004, label smoothing of 0.1, dropout of 0.4 with no attention or activation dropout, and 40k characters per batch, for 50 epochs.

All MTL models have 6 encoder and decoder layers, encoder and decoder embedding dimensions of 1280, encoder and decoder feed-forward network embedding dimensions of 12288, and 20 heads. All embeddings are shared. We train with a learning rate of 0.0004, label smoothing of 0.1, dropout of 0.2 with attention and activation dropout each set to 0.1, and 40k characters per batch, for 20 epochs. We select the best performing model for each run using dev WER.

## 4 Results

The word error rates for our method (main model, both ablation models, and baseline) are shown in Table 5, along with error rates reported by prior work. Our main model achieves 4.71% WER on the development set, a relative improvement of 22.8% over the previous best development set result from Zalmout and Habash (2020), who trained a multitask model on PATB features and incorporated a morphological analyzer. On the test set, it achieves 4.79% WER, a relative improvement of 18.8% over the best previous test set result from Qin et al. (2021), who trained a BERT-based model.[11]

Our ablation models also outperform all prior work, with the model trained on Ar$\rightarrow *$ bitext outperforming the model trained on $*\rightarrow$Ar bitext, but neither perform as well as the main model trained on both Ar$\rightarrow *$ and $*\rightarrow$Ar. See §5 for more detailed comparisons between the models trained in this work.

Finally, our baseline model (a character-based Transformer trained without bitext) slightly outperforms to prior models from Alqahtani et al. (2019b) and Alqahtani and Diab (2019), that also do not use MTL, morphological analyzers, or contextual embeddings.

## 5 Automatic Analysis

### 5.1 Case Endings

We compute the Diacritic Error Rate (DER) for all models trained in this work for several different settings: all characters (including whitespace, punctuation, and non-Arabic characters), Arabic characters, Arabic case endings, and Arabic characters excluding case endings: see Table 6. We use POS tags to determine which words in the test set have case endings.[12] Comparing our main model to the baseline, we see that MTL improves case endings more than non-case endings: case ending DER is improved by 47.7% (3.61% vs 6.90%) as compared to 30.2% (1.73% vs 2.48%) for non case ending characters. Furthermore, comparing the ablation models, the performance difference between them is more pronounced on case endings, where the ∗→Ar model is 12.7% behind the Ar→∗ model, while the difference is only 5.1% for non case endings.

[11] We exclude the experiments of Qin et al. (2021) which use Farasa in training, as Farasa was trained on the test set (see §2.1.1).

[12] Several prior works have reported DER of just the last character as a stand-in for case-ending DER However, this analysis is muddied by the fact that not all words in Arabic have case endings; in the PATB test set, for example, the POS tags indicate that only about 46.8% of words have them.

| | Multitask? | Morphological Analyzer? | Word Embeddings? | Dev WER ↓ | Test WER ↓ |
|---|---|---|---|---|---|
| Alqahtani et al. (2019b) | No | No | No | | 8.20% |
| Alqahtani and Diab (2019) | No | No | No | | 7.60% |
| Alqahtani et al. (2020) | PATB Features | No | fastText | | 7.51% |
| Zalmout and Habash (2019) | PATB Features | Train & Test | fastText | 7.30% | 7.50% |
| Zalmout and Habash (2020) | PATB Features | Train & Test | fastText | 6.10% | |
| Qin et al. (2021)† | No | No | Zen 2.0 | 6.49% | 5.90%‡ |
| This word (baseline) | No | No | No | 7.46% | 7.49% |
| This work (ablation) | Translate ∗→Ar | No | No | 5.60% | 5.83% |
| This work (ablation) | Translate Ar→∗ | No | No | 5.24% | 5.32% |
| **This work** | Translate ∗→Ar & Ar→∗ | No | No | **4.71%** | **4.79%** |

Table 5: Development and Test WER (lower is better) for our main system, ablation systems, and baseline, compared to recent work. Our main system outperforms all prior work, as do both ablation systems. †:We exclude the experiments of Qin et al. (2021) which use Farasa in training, as Farasa was trained on the test set (see §2.1.1). ‡:Mean of 5 runs with different random seeds.

| | Baseline | Multitask Learning ∗→Ar | Ar→∗ | Both |
|---|---|---|---|---|
| All | 2.34% | 1.85% | 1.73% | **1.52%** |
| Arabic | 2.97% | 2.35% | 2.21% | **1.94%** |
| Arabic CE | 6.90% | 4.71% | 4.18% | **3.61%** |
| Arabic non-CE | 2.48% | 2.06% | 1.96% | **1.73%** |

Table 6: Diacritic error rate for all characters (including whitespace and non-Arabic characters), Arabic characters only, Arabic case endings (CE), and Arabic characters excluding case endings (non-CE). We use POS tags to determine which words contain case endings.

## 5.2 WER vs Sentence Length

We show WER as a function of sentence length (in undiacritized characters) in Figure 2. We note that while both the ∗→Ar and Ar→∗ models tend to improve with sentence length, the improvement is much more pronounced for the Ar→∗ model. In other words, the Ar→∗ model is benefiting much more from increased context than the ∗→Ar model.

In conjunction with the DER results in §5.1, this indicates that training the model to translate out of Arabic is more helpful at injecting semantic and linguistic knowledge into the model to address ambiguity. The fact that the two translation directions are complementary suggests that training the model to translate into Arabic is addressing data sparsity issues in the model's decoder, despite the mismatch between the bitext being undiacritized and the model needing to produce diacritized output.

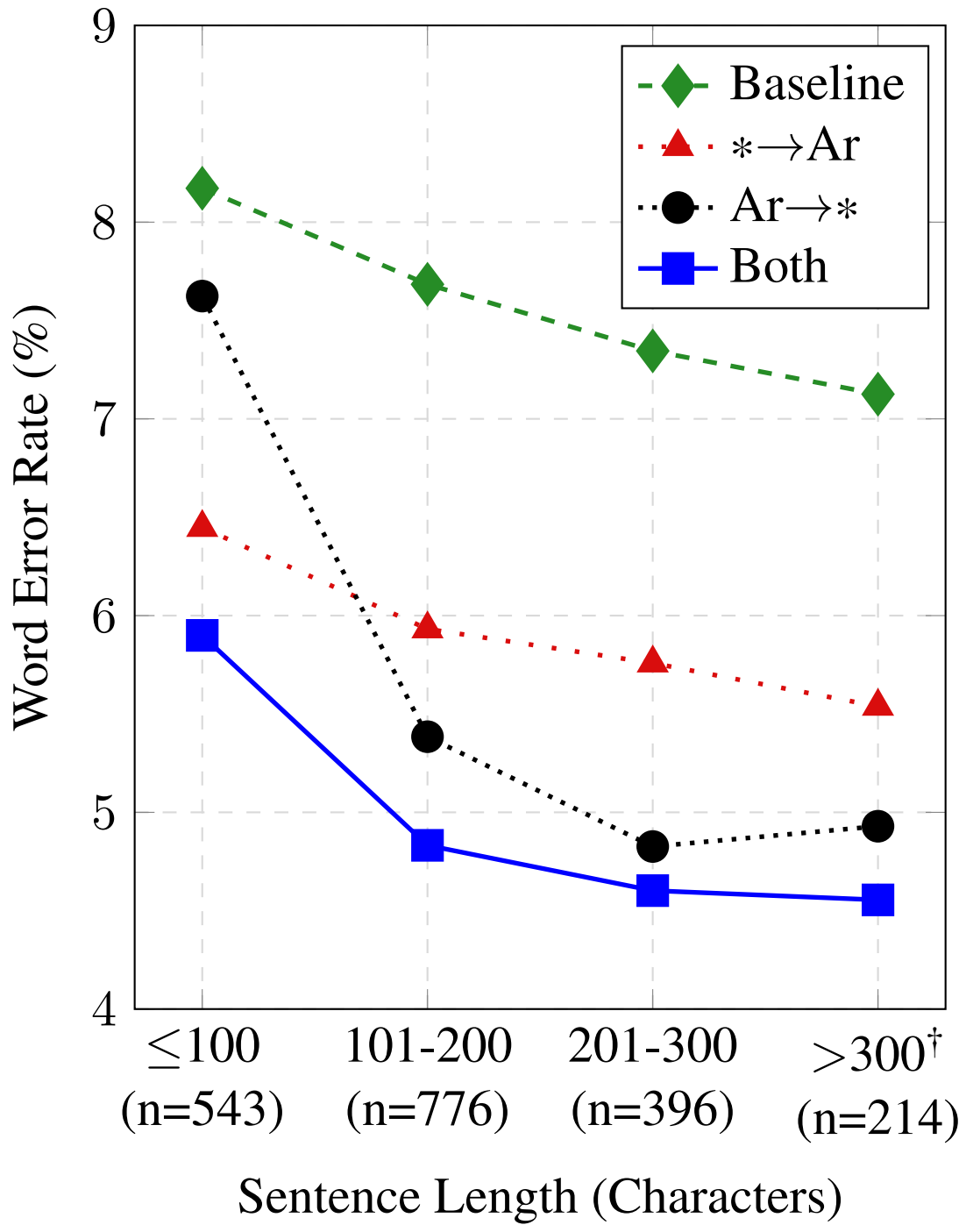

Figure 2: Word error rate vs (undiacritized) character length. †:Sentences over 300 characters long are processed in overlapping windows of 300 characters (see §3.4).

| | Male | | Female | | Bias |
|---|---|---|---|---|---|
| | Count | WER | Count | WER | |
| Pronoun | 835 | 6.23% | 641 | 8.11% | 30.3% |
| Verb | 3579 | 5.34% | 2083 | 6.39% | 19.6% |
| Suffix | 901† | 5.22% | 10222 | 5.71% | 9.5% |

Table 7: WER for male vs female pronouns, verbs, and nouns/adjectives with gendered suffixes. †: We include only suffixes which are explicitly marked in the PATB for gender, which tend to be female: see §5.3

.

### 5.3 Gender Bias

Gender bias has been noted in many aspects of NLP (Sun et al., 2019) but we are not aware of any prior work looking at gender bias in diacritization. We use the PATB POS tags to isolate three types of gendered words: pronouns, verbs, and suffixes. We use the term "suffix" to refer to nouns and adjectives that have a gendered suffix. Unsurprisingly, we find that the model is better at diacritizing male words than female words in all three cases (see Table 7), with words in the female categories being diacritized incorrectly 9.5%-30.3% more often than for their male equivalents.

We suspect that this bias is due at least in part to representation within the data: Male pronouns and verbs are 30% and 72% more common than their female counterparts, respectively. Counts of suffixes are complicated by the fact that that PATB only marks certain nouns and adjectives for gender (including those with taa marbuta (ة), which tend to be female). By manual inspection, the remainder appear to be male; however, we were unable to confirm this in the annotation guidelines so we included only suffixes explicitly marked for gender.

### 5.4 WER vs POS

The PATB includes detailed POS tagging. We exploit this feature to examine how our model performs on different parts of speech: see Table 8. Note that the PATB has one or more POS tags per word, with about 2.19 tags per word on average in the test set. We do not attempt to split words into their respective parts, as we find cases where this is not straightforward. Instead, such words are counted multiple times. As an example, الأَوَّلُون (the first) is both a determiner and cardinal adjective, and contributes to the WER of both.

For the parts of speech we consider that have at least 500 occurrences in the test set, the worst performing POS for the MTL model by far is proper nouns (count=5969) at 14.09% WER. This is followed by imperfect verbs (count=2598) at 7.89% WER, possessive pronouns (count=1609) at 6.60%, and adjectives (excluding cardinal and comparative) (count=6106) at 6.49%.

Comparative adjectives, which are relatively infrequent (count=264) also have a high WER of 9.95%, but the worst POS considered by far is the extremely infrequent (count=18) imperative verbs, with a WER of 72.22%. Imperative verbs illustrate the importance of domain; news data contains very few imperatives, and imperative verbs (like most words in Arabic) are often distinguished from other words by diacritics alone. For example, imperatives can often have the same form as perfective verbs as in استمر , which can be imperative as in اِستَمِر (continue) or اِستَمَرَّ (he continued). We confirm via manual inspection that the majority of the imperative verb errors are in fact homograph errors.

### 5.5 WER vs Word Frequency

MTL improves learning across all word frequencies: see Table 9. The biggest improvements are seen for words seen once and 2-4 times in training, with relative improvements of 43.6% and 45.5%, respectively.

## 6 Manual Analysis

We manually annotate all differences between our model prediction and the gold test set for a randomly selected 20% of the 1246 sentences in the test set that contain at least one disagreement.

We find that approximately 66% of the disagreements between the gold test set and the model are the result of model errors, which we denote as "true errors." The majority of these errors are due to case markings being either incorrect (38.6% of all true errors) or missing (16.5% of all true errors), while the rest of the word is correct.

However, we find that the model is actually correct in approximately 34% of disagreements between the model output and the gold refer-ence output. We denote such cases as "false errors." Nearly half of the false errors were due to the test set missing diacritics (47.7% of all false errors) and another 29.6% of all false errors were due to errors in the test set diacritics. Finally, 5.1% of false errors are cases where the input to the model is not a real word, making the correct output undefined. A very small number of words (3.2% of false errors) had trivial diacritic variations (e.g. one having a sukun while the other had no diacritic, or one having a fatha before an alif while the other did not). 10.2% of the false errors were the result of valid variations which did not change the meaning of the sentence in any way (e.g. يُكْشِفُ vs يَكْشِفُ and الدُّوَلِي vs الدَّوْلِي). Finally, only 4.2% of false errors were the result of valid variations that changed the meaning of the sentence in some way while still resulting in a plausible meaning. Our manual analysis suggests that the true WER of our MTL model is approximately 3.15%.

|  | Count | Baseline WER | MTL WER | Rel. imprv. | Examples |
|---|---|---|---|---|---|
| Noun: Proper | 5969 | 18.24% | **14.09%** | 22.8% | مَريَم (Mary); أَحمَد (Ahmed) |
| Noun: Numeric | 1609 | 3.29% | **2.11%** | 35.8% | عَشَرَة (ten); أَربَعَة (four) |
| Noun: Quantity | 451 | 10.42% | **5.32%** | 48.9% | أَيَّة (any; fem); بَعض (some) |
| Noun: Other | 22795 | 8.43% | **5.03%** | 40.3% | يَوم (day); دُوَيلَة (small country) |
| Pronoun: Possessive | 1681 | 11.42% | **6.60%** | 42.2% | كِتابِيَ (my book); كِتابُكُن(your book; fem) |
| Pronoun: Demonstrative | 601 | **0.00%** | 0.17% | - | هٰذا (this; male singular); هاتان (these, fem dual) |
| Pronoun: Other | 1154 | 1.04% | **0.52%** | 50.0% | شاهَدَتنِي (she saw me); أَنتَ (you; male singular) |
| Verb: Inflected, Perfect | 3273 | 9.53% | **4.89%** | 48.7% | ذَهَبَ (he went); قُبِلَ (it was accepted) |
| Verb: Inflected, Imperfect | 2598 | 13.55% | **7.89%** | 41.8% | يَذهَبُ (he goes); تُقبَلُ (it is accepted) |
| Verb: Inflected, Imperative | 18 | 83.33% | **72.22%** | 13.3% | اِذهَب (go; male); قِفِي (stop; fem) |
| Adverb | 260 | **0.00%** | 0.38% | - | مَتَى (when); حِينَذاك (then) |
| Adjective: Cardinal | 348 | 7.18% | **4.31%** | 40.0% | القَرن (19th century); الأَوَّلُون (the first) |
| Adjective: Comparative | 264 | 16.67% | **9.85%** | 40.9% | أَحرَصُ (more cautious); الأَحسَن (the best) |
| Adjective: Other | 6106 | 10.87% | **6.49%** | 40.4% | تارِخِيٌّ (historic); يَهُودِيٌّ (Jewish) |
| Determiner | 15337 | 8.72% | **5.85%** | 32.9% | التُونِسِي (the Tunisian); اليَومُ (the day) |

Table 8: WER for our baseline and our main MTL model, for various parts of speech, and their associated count in the test set. Note: many words have more than one POS and contribute to 2+ categories (see §5.4).

| # Occurrences in PATB-train | Baseline | Multitask Learning ∗→Ar | Ar→∗ | Both |
|---|---|---|---|---|
| 0 | 30.93% | 26.30% | 23.20% | **21.92%** |
| 1 | 17.63% | 12.46% | 10.33% | **9.95%** |
| 2-4 | 11.94% | 8.32% | 7.56% | **6.51%** |
| 5-16 | 8.78% | 6.83% | 6.50% | **5.67%** |
| 17-64 | 7.80% | 5.81% | 5.50% | **4.86%** |
| 65-256 | 6.33% | 4.97% | 4.55% | **3.76%** |
| 257-1024 | 4.34% | 3.28% | 3.16% | **2.94%** |
| >1024 | 0.30% | **0.20%** | 0.29% | 0.22% |

Table 9: WER vs number of times a word occurs in PATB-train (ignoring diacritics), for all four models trained in this work.

## 7 Conclusion

We demonstrate that training a diacritics model to both diacritize and translate substantially outperforms a model trained on the diacritization task alone. Adding translation data substantially increases the amount of training data seen by the model, addressing data sparsity issues in diacritization. The translation task also injects semantic and linguistic knowledge into the model, helping the model resolve ambiguities in diacritization.

Our method achieves a new state-of-the-art word error rate of 4.79% on the Penn Arabic Treebank datasets, using the standard data splits of Diab et al. (2013). Manual analysis of the errors indicate the true error rate is actually around 3.15%, after accounting for errors in the test set and plausible variations between the model output and test set.

Finally, our manual and automatic analysis points to several remaining challenges in Arabic diacritization, including proper nouns, female word forms, and case endings.